\documentclass[letterpaper,journal]{IEEEtran}
\ifdefined\pdfminorversion
\fi

\usepackage{amsmath,amssymb,amsfonts}
\usepackage{array}
\usepackage[caption=false,font=normalsize,labelfont=sf,textfont=sf]{subfig}
\usepackage{textcomp}
\usepackage{stfloats}
\usepackage{url}
\usepackage{graphicx}
\usepackage{cite}
\usepackage{bm}
\usepackage{booktabs}
\usepackage[table]{xcolor}
\usepackage{tabularx}

\usepackage[hidelinks]{hyperref}
\usepackage[capitalize]{cleveref}
\crefname{figure}{Fig.}{Figs.}
\Crefname{figure}{Fig.}{Figs.}
\crefname{table}{Table}{Tables}
\Crefname{table}{Table}{Tables}
\crefname{section}{Section}{Sections}
\Crefname{section}{Section}{Sections}
\crefname{appendix}{Appendix}{Appendixes}
\Crefname{appendix}{Appendix}{Appendixes}

\newcommand{\vs}{s}
\newcommand{\vt}{t}
\newcommand{\Vsup}{\mathcal{V}_{\mathrm{sup}}}

\newcommand{\Iaif}[1][]{\mathbf{I}^{#1}}
\newcommand{\Idef}[1][]{\mathbf{I}_{\mathrm{def}}^{#1}}
\newcommand{\Kdef}[1][]{k_{\mathrm{def}}^{#1}}
\newcommand{\Irender}[1][]{\hat{\mathbf{I}}^{#1}}
\newcommand{\Gauss}{\mathcal{G}}

\newcommand{\depth}[1][]{\mathbf{Z}^{#1}}
\newcommand{\disp}[1][]{\mathbf{D}^{#1}}
\newcommand{\depthfinal}[1][]{\mathbf{Z}_{\mathrm{ref}}^{#1}}
\newcommand{\dispfinal}[1][]{\mathbf{D}_{\mathrm{ref}}^{#1}}
\newcommand{\depthrender}[1][]{\hat{\mathbf{Z}}^{#1}}
\newcommand{\disprender}[1][]{\hat{\mathbf{D}}^{#1}}
\newcommand{\dispda}[1][]{\mathbf{D}_{\mathrm{DA}}^{#1}}
\newcommand{\Fda}{\mathbf{F}_{\mathrm{DA}}}
\newcommand{\dispunet}[1][]{\mathbf{D}_{\mathrm{U}}^{#1}}
\newcommand{\deltadispunet}{\Delta\dispunet}
\newcommand{\confunet}[1][]{\mathbf{C}_{\mathrm{U}}^{#1}}
\newcommand{\dispalign}[1][]{\mathbf{D}_{\mathrm{align}}^{#1}}
\newcommand{\deltadispalign}{\Delta\dispalign}
\newcommand{\scale}{a}
\newcommand{\normscalemap}{\mathbf{S}}
\newcommand{\scalerange}{\rho}

\newcommand{\px}{\mathbf{p}}
\newcommand{\pxset}{\Omega}
\newcommand{\pxvalidset}{\Omega_{\mathrm{valid}}}
\newcommand{\coc}{r}
\newcommand{\depthfocus}[1][]{z_{\mathrm{f}}^{#1}}
\newcommand{\dispfocus}[1][]{d_{\mathrm{f}}^{#1}}
\newcommand{\deltadisp}{\Delta\disp}
\newcommand{\deltadispmax}{{\Delta d}_{\max}}

\newcommand{\dmin}[1][]{\min_{\px \in \pxvalidset}(\disp[#1](\px))}
\newcommand{\dmax}[1][]{\max_{\px \in \pxvalidset}(\disp[#1](\px))}
\newcommand{\dispnorm}[1][]{\mathbf{D}'^{\,#1}}
\newcommand{\dispnormrender}[1][]{\hat{\mathbf{D}}'^{\,#1}}
\newcommand{\dispnormunet}[1][]{\mathbf{D}_{\mathrm{U}}'^{\,#1}}

\newcommand{\unetdefocus}{f_{\mathrm{U1}}}
\newcommand{\unetrefine}{f_{\mathrm{U2}}}

\newcommand{\papertitle}{DfD-GS: Defocus Image Prior for Metric-Scale Single-View Depth Estimation and View Synthesis}
\hypersetup{
  pdftitle={\papertitle},
  pdfauthor={Shiori Ueda, Rinto Yagawa, Hideo Saito, Shohei Mori},
  pdflang={en}
}

\begin{document}
\pagestyle{plain}

\title{\papertitle}
\author{
  Shiori Ueda\textsuperscript{1}\quad
  Rinto Yagawa\textsuperscript{1}\quad
  Hideo Saito\textsuperscript{1}\quad
  Shohei Mori\textsuperscript{2,1} 
  
  \textsuperscript{1}\,Keio University\quad
  \textsuperscript{2}\,University of Stuttgart%

\thanks{This work was supported by JSPS KAKENHI Grant Number JP24KJ1962, the Alexander von Humboldt Foundation funded by the German Federal Ministry of Research, Technology and Space, and the Deutsche Forschungsgemeinschaft (DFG, German Research Foundation) under Germany's Excellence Strategy -- EXC 2120/2 -- 390831618.}}

\maketitle
\thispagestyle{plain}

\begin{abstract}
  Recovering metric-scale geometry from observations at a single viewpoint remains fundamentally ill-posed, limiting the reliability of novel view synthesis. Recent methods rely on learned monocular depth priors, which provide strong structural estimates but can suffer from scale ambiguity and dataset-dependent biases.
  We propose DfD-GS, a feed-forward framework that predicts a metric-scale 3D Gaussian scene representation from a single-view all-in-focus--defocused image pair. The key idea is to exploit defocus blur as a physical metric depth cue.
  DfD-GS first estimates a defocus-based metric disparity map from the all-in-focus--defocused image pair and the corresponding defocus strength. This estimate is used to align and refine a monocular depth prior to metric scale. Visible and hidden 3D Gaussians are then predicted for occlusion-aware novel view synthesis.
  Experiments on synthetic-defocus datasets constructed from Booster and Middlebury show that DfD-GS outperforms single-image baselines in metric depth estimation and novel view synthesis. On WildRGB-D, DfD-GS achieves lower AbsRel and LPIPS but higher depth RMSE than the single-image baselines. Additional analyses show that DfD-GS maintains a more stable metric scale under field-of-view changes. Qualitative results on a real dataset show recovery of the main scene structure under optical defocus.
  Code is available on our project page: \url{https://sh1027.github.io/dfd-gs/}.
\end{abstract}

%-------------------------------------------------------------------------
\section{Introduction}
\label{sec:introduction}

%1. What is the problem, why is it hard?
\IEEEPARstart{S}{ingle-view} 3D reconstruction and novel view synthesis aim to infer the geometry and appearance of a scene from a single viewpoint. This problem is fundamental to image-based rendering and immersive applications such as augmented reality, virtual reality, and digital twins~\cite{zhang2026advances}, where only limited observations are available.
Recent methods have addressed single-view novel view synthesis by lifting a monocular image into 3D scene representations, such as layered representations~\cite{tucker2020single-view,han2022adampi} and 3D Gaussians~\cite{szymanowicz2025flash3d,mescheder2026sharp}.
However, recovering a metrically accurate 3D scene from a single image remains ill-posed because the same image can be explained by multiple scene scales and depth configurations.

\begin{figure}[t]
  \centering
  \includegraphics[width=\linewidth]{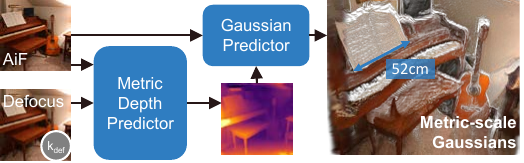}
  \caption{Given a single-view all-in-focus--defocused image pair and the corresponding defocus strength, DfD-GS predicts metric-scale 3D Gaussians by exploiting defocus blur as a physical depth cue.}
  \label{fig:teaser}
\end{figure}

%2. How is it solved today?
A typical strategy is to rely on monocular metric depth priors learned from large-scale datasets. Recent monocular metric depth foundation models provide strong geometric estimates and generalize across diverse scenes~\cite{hu2024metric3dv2,piccinelli2025unidepthv2, bochkovskii2025depthpro}.
Nevertheless, metric depth estimation from a single RGB image remains fundamentally under-constrained. Camera parameters can reduce projection ambiguity, but absolute scale and local geometry must still be inferred from learned, data-driven priors, which can become unreliable when the scene content, object scale, or imaging condition deviates from the training distribution. Such geometric errors can degrade novel view synthesis, where inaccurate metric geometry leads to misprojection and distorted scene structure.

Defocus blur provides a complementary cue for metric 3D reconstruction.
Depth from defocus (DfD) estimates scene depth from optical blur, whose magnitude is physically related to the distance from the focus plane~\cite{pentland1987new}.
This image-formation cue is directly tied to metric geometry, but it is also inherently ambiguous. Blur estimation is texture-dependent and less reliable in flat or saturated regions. This motivates a hybrid approach that combines the global structural prior of monocular depth foundation models with the physical metric cue provided by defocus blur.

%3. What is the new technical idea; why can we succeed now?
We propose DfD-GS, a feed-forward framework for metric-scale single-view novel view synthesis from a single-view all-in-focus--defocused image pair (\cref{fig:teaser}). Our key idea is to combine the global structural prior of a monocular depth foundation model with local metric cues from defocus blur. DfD-GS estimates a defocus-based metric disparity map from the all-in-focus--defocused image pair and the corresponding defocus strength and uses it to align and refine the monocular depth prior to metric scale. It then lifts the refined source-view geometry into an occlusion-aware 3D Gaussian representation that separately models visible and hidden scene content for novel view rendering. This unified formulation enables both metric depth estimation and single-view novel view synthesis in a single forward pass.

%4. What is the impact if successful?
To evaluate metric depth estimation and novel view synthesis, we construct synthetic-defocus datasets from RGB-D datasets with metric depth. DfD-GS outperforms the single-image baselines on Booster~\cite{ramirez2024booster} and Middlebury~\cite{scharstein2014middlebury} for both tasks. On WildRGB-D~\cite{xia2024wildrgbd}, it achieves the lowest AbsRel and LPIPS, but has higher depth RMSE than the single-image baselines. We also analyze the effect of aperture size and show that DfD-GS maintains a more stable metric scale under field-of-view changes. Evaluation on a real dataset further shows that the method recovers the main scene structure from images captured with optical defocus.

Our main contributions are as follows:
\begin{itemize}
  \item We propose DfD-GS, a feed-forward framework that predicts a metric-scale 3D Gaussian scene representation from a single-view all-in-focus--defocused image pair for depth estimation and novel view synthesis.
  \item We introduce a defocus-guided metric depth prediction module that uses defocus-based metric disparity estimates to globally align and locally refine a monocular depth prior.
\end{itemize}

%-------------------------------------------------------------------------
\section{Related Work}
\label{sec:relatedWork}

\subsection{Single-view Novel View Synthesis}
Single-view novel view synthesis aims to render unseen viewpoints from a single viewpoint.
A common approach is to lift the input image into an explicit 3D-aware representation, such as layered depth images or multiplane images, and render target views from the reconstructed scene~\cite{zhou2018stereo,shih20203dphoto,tucker2020single-view,jampani2021slide,han2022adampi}.

Recently, 3D Gaussian Splatting~\cite{kerbl20233dgs} has motivated feed-forward Gaussian prediction from single images~\cite{szymanowicz2024splatter,szymanowicz2025flash3d,mescheder2026sharp}.
Splatter Image~\cite{szymanowicz2024splatter} predicts object-level Gaussians from a single image, while Flash3D~\cite{szymanowicz2025flash3d} extends monocular depth priors to feed-forward scene reconstruction using layered 3D Gaussians.
Although these methods achieve efficient single-view reconstruction and view synthesis, metric scale is not explicitly considered.
SHARP~\cite{mescheder2026sharp} explicitly targets high-resolution photorealistic view synthesis from a metric 3D Gaussian representation. However, its metric scale is still inferred from a single RGB image in a data-driven manner.
In contrast, our method does not infer metric scale solely from RGB appearance. It exploits defocus blur as a physical cue to estimate metric-scale depth and construct a metric-scale 3D Gaussian representation.

\subsection{Metric Monocular Depth Estimation}
Monocular depth estimation seeks to infer dense scene geometry from a single RGB image. Large-scale relative depth models have improved cross-dataset generalization by learning robust monocular depth priors~\cite{ranftl2022midas,yang2024dav2}.
Metric monocular depth estimation further aims to recover depth in absolute scale. Early zero-shot metric methods addressed scale-aware prediction by learning geometric scale priors~\cite{guizilini2023towards} or combining relative and metric depth supervision~\cite{bhat2023zoedepth}.

Recent foundation models improve zero-shot metric depth by explicitly modeling the relationship between depth and camera properties in different ways. Metric3D~\cite{yin2023metric3d} and Metric3D v2~\cite{hu2024metric3dv2} transform images and depths into canonical camera spaces, Depth Pro~\cite{bochkovskii2025depthpro} predicts canonical inverse depth with focal length estimation, and Depth Anything 3~\cite{lin2026da3} introduces a metric model that predicts canonical metric depth. UniDepth~\cite{piccinelli2024unidepth} and UniDepth v2~\cite{piccinelli2025unidepthv2} estimate dense camera representations with pseudo-spherical outputs. MoGe~\cite{wang2025moge} predicts affine-invariant point maps for open-domain 3D geometry.
Although these methods reduce ambiguity arising from camera parameters, absolute depth remains inherently ambiguous from a single RGB image and is therefore inferred largely from learned data-driven priors.
In contrast, our method complements the monocular prior with defocus blur as a physically grounded metric cue.

\subsection{Depth Estimation from Defocus Cues}
Defocus cues have long been used for depth estimation through depth from focus (DfF) and depth from defocus (DfD).
DfF estimates depth from a focal stack captured with different focus settings. Classical DfF methods infer the depth at which each scene point appears sharp using local focus measures~\cite{nayar1994shape,pertuz2013analysis,suwajanakorn2015depth}.
Learning-based methods replace hand-crafted focus measures with neural networks and estimate depth from focal stacks. Early methods relied on direct depth supervision~\cite{hazirbas2019ddff}, while later approaches introduced defocus-based intermediate supervision~\cite{maximov2020defocusnet}, all-in-focus image supervision~\cite{wang2021bridging}, differential focus-volume representations~\cite{yang2022dffv}, self-supervised learning from sparse focal stacks~\cite{si2023dered}, and defocus models for camera-setting invariance~\cite{fujimura2023deep}. HybridDepth~\cite{ganj2025hybriddepth} further combines DfF cues from focal stacks with single-image depth priors to produce robust metric depth maps. These methods exploit rich focus variation across multiple images, but they generally require focal stacks, which increase capture cost and computational memory.

DfD, in contrast, does not require a dense focal stack and estimates depth from defocus blur in one or a few images. Classical methods established this idea by relating defocus blur to scene depth using analytical image-formation models, spatial-domain transforms, or filter-based relative defocus measurements~\cite{pentland1987new,subbarao1994depth,watanabe1998rational}. Recent learning-based methods have used defocus either as a supervisory cue for monocular depth learning~\cite{gur2019single} or as an input cue for joint depth estimation and focused-image restoration~\cite{nazir2023depth}. These methods demonstrate that defocus provides useful geometric information with limited inputs.
Inspired by DfD, our method uses a single-view all-in-focus--defocused image pair to refine a monocular depth prior and construct a metric-scale 3D Gaussian representation for novel view synthesis.

%-------------------------------------------------------------------------
\section{Method}
\label{sec:method}

\begin{figure*}[t]
  \centering
  \includegraphics[width=\linewidth]{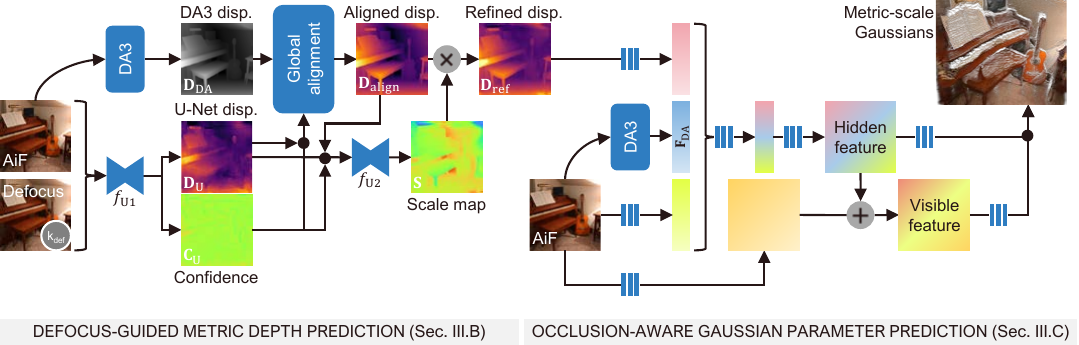}
  \caption{Overview of DfD-GS. A defocus-conditioned U-Net estimates a metric disparity offset from the all-in-focus--defocused image pair and the defocus strength. This estimate is used to globally align the normalized disparity-offset prior from DA3. The aligned offset is converted to metric disparity and refined using a predicted scale map. The refined depth, all-in-focus image, and DA3 features are then used to predict visible and hidden 3D Gaussians for novel view synthesis.}
  \label{fig:pipeline}
\end{figure*}

\begin{table}[t]
  \caption{Main notation used in the proposed method.}
  \label{tab:notation}
  \centering
  \begin{tabularx}{\linewidth}{@{}lX@{}}
    \toprule
    Symbol & Definition \\
    \midrule
    $\Iaif$, $\Idef$, $\Kdef$ & All-in-focus image, defocused image, and defocus strength. \\
    $\depthfocus$, $\dispfocus$ & Known focus distance and its inverse ($\dispfocus=1/\depthfocus$). \\
    $\depth$, $\disp$ & Metric depth and disparity ($\disp=1/\depth$). \\
    $\deltadisp$ & Disparity offset from the focus plane: $\deltadisp=\dispfocus-\disp$. \\
    $\Fda$ & Multi-scale DA3 feature maps. \\
    $\dispda$ & DA3 normalized disparity-offset prior. \\
    $\deltadispunet$, $\dispunet$ & U-Net disparity offset and derived metric disparity. \\
    $\confunet$ & Confidence map for the U-Net offset estimate. \\
    $\deltadispalign$, $\dispalign$ & Globally aligned disparity offset and derived metric disparity. \\
    $\normscalemap$ & Predicted scale map for dense disparity refinement. \\
    $\depthfinal$, $\dispfinal$ & Refined source-view metric depth and disparity. \\
    $\Gauss$ & Predicted set of visible and hidden 3D Gaussians. \\
    $\Irender[v]$ & Rendered RGB image at view $v$. \\
    $\depthrender[v]$, $\disprender[v]$ & Rendered metric depth and disparity at view $v$. \\
    \bottomrule
  \end{tabularx}
\end{table}
%-------------------------------------------------------------------------

We introduce DfD-GS, a feed-forward method for predicting a metric-scale 3D Gaussian scene representation from a single-view all-in-focus--defocused image pair. As shown in \cref{fig:pipeline}, the input consists of an all-in-focus image $\Iaif$, a defocused image $\Idef$, and the corresponding defocus strength $\Kdef$. We additionally assume that the focus distance $\depthfocus$ is known from the camera configuration. In this work, the focus plane is placed at the nearest scene depth. Our objective is to estimate a metric-scale 3D Gaussian representation $\Gauss$ that can synthesize novel views.
The main notation used in this section is summarized in \cref{tab:notation}.

Our method consists of three stages: (1) DA3-based monocular depth prior extraction, (2) defocus-guided metric depth prediction, and (3) occlusion-aware Gaussian parameter prediction (\Cref{fig:pipeline}).

\subsection{DA3-based Monocular Depth Prior Extraction}
\label{sec:method_da3}

We extract multi-scale image features $\Fda$ and a depth map from $\Iaif$ using the frozen Depth~Anything~3 (DA3)~\cite{lin2026da3} backbone.
We convert the depth map to disparity and linearly rescale it to $[0,1]$, mapping the nearest and farthest predicted points to $0$ and $1$, respectively. The resulting normalized disparity-offset map $\dispda$ serves as a relative-geometry prior for defocus-guided metric alignment in the next stage.

\subsection{Defocus-guided Metric Depth Prediction}
\label{sec:method_metricdepth}

Defocus provides a metric depth cue through the physical relationship between blur radius and scene depth. We use this cue to globally align and locally refine the DA3 prior $\dispda$.

For each pixel $\px$ in the image grid $\pxset$, we define metric disparity as inverse depth, $\disp(\px)=1/\depth(\px)$, measured in $\mathrm{m}^{-1}$, and let $\dispfocus=1/\depthfocus$ denote the focus disparity. Under the nearest-focus assumption, the disparity offset $\deltadisp(\px)=\dispfocus-\disp(\px)$ is nonnegative. The blur radius in pixels is modeled as
\begin{equation}
  \coc(\px)
  = \Kdef\left|\disp(\px)-\dispfocus\right|
  = \Kdef\,\deltadisp(\px),
  \label{eq:defocus_relation}
\end{equation}
where $\Kdef$ is a scalar parameter controlling defocus strength. We compute it from known camera parameters, as detailed in the supplementary document (\cref{sec:defocus_strength}).

\subsubsection{Defocus-conditioned U-Net}

We use a U-Net $\unetdefocus$ that takes the image pair $\{\Iaif,\Idef\}$ as input and is conditioned on $\Kdef$. The network has two sigmoid-activated branches for estimating a metric disparity offset and its confidence. We obtain the nonnegative metric disparity offset as
\begin{equation}
  \deltadispunet(\px)
  = \deltadispmax\left(1-u(\px)\right),
  \label{eq:unet_disparity_offset}
\end{equation}
where $u(\px)\in[0,1]$ is the offset branch output and $\deltadispmax$ is a predefined upper bound on the predicted metric disparity offset. The other branch predicts $\confunet(\px)\in[0,1]$, representing the reliability of the offset estimate.
For depth supervision, we obtain $\dispunet$ by applying a smooth positive floor to $\dispfocus-\deltadispunet$ (supplementary document, \cref{eq:unet_metric_disp}).

\subsubsection{Global alignment to the defocus estimate}

With the focus plane fixing the offset origin, we align $\dispda$ to $\deltadispunet$ using a single global scale $\scale$. We estimate $\scale$ by confidence-weighted least squares:
\begin{equation}
  \min_{\scale}
  \sum_{\px\in\pxset}
  \confunet(\px)
  \left[
    \scale\dispda(\px)-\deltadispunet(\px)
  \right]^2.
  \label{eq:global_alignment_objective}
\end{equation}
Confidence weighting reduces the influence of unreliable defocus estimates.
The aligned disparity offset is
\begin{equation}
  \deltadispalign(\px)=\scale\dispda(\px).
  \label{eq:aligned_disparity_offset}
\end{equation}
We obtain the aligned metric disparity $\dispalign$ from $\dispfocus-\deltadispalign$ with a positive lower bound (supplementary document, \cref{eq:aligned_metric_disparity}).

\subsubsection{Dense disparity refinement}

Global alignment sets the overall metric scale but cannot correct local errors in the monocular depth prior~\cite{ganj2025hybriddepth}.
A second U-Net, $\unetrefine$, takes $1-\deltadispalign/\deltadispmax$, $1-\deltadispunet/\deltadispmax$, and $\confunet$ as input and predicts a scale map $\normscalemap\in[-1,1]$. The normalized offset values are 1 at the focus plane and decrease with increasing disparity offset.

We set the correction range to $\scalerange=\log(q_{90}/q_{10})$, where $q_{10}$ and $q_{90}$ are the 10th and 90th percentiles of $\dispalign$ over image pixels. These percentiles limit the influence of extreme disparities. The positive multiplicative correction $\exp(\scalerange\normscalemap)$ lies in $[q_{10}/q_{90},\,q_{90}/q_{10}]$.
We obtain $\dispfinal$ from $\dispalign\exp(\scalerange\normscalemap)$, enforcing positive disparity and the nearest-focus assumption (supplementary document, \cref{eq:refined_metric_disparity}). The refined metric depth is $\depthfinal(\px)=1/\dispfinal(\px)$.

\subsection{Occlusion-aware Gaussian Parameter Prediction}
\label{sec:method_gaussians}

To handle disocclusions in nearby novel views, we predict two 3D Gaussians for each source-view pixel: a visible Gaussian for the observed surface and a hidden Gaussian for scene content that may become visible from nearby viewpoints.
For an input image of size $H \times W$, the predicted Gaussian set contains $2HW$ Gaussians and is denoted by $\Gauss$.

\subsubsection{Feature fusion and parameter prediction}

To predict Gaussian parameters, we fuse RGB image-token features, multi-scale DA3 features $\Fda$, and depth-token features.
$\Iaif$ and $\depthfinal$ are first embedded into patch tokens.
They are combined with $\Fda$ and fused by an MLP to produce a dense feature map at the input resolution.
The hidden Gaussian parameters are predicted from the fused feature map. For the visible Gaussian, we additionally use local RGB cues from $\Iaif$, since its appearance is directly supported by the source-view observation.

\subsubsection{Lifting to 3D and rendering}

Each predicted Gaussian is lifted to 3D using the refined metric depth and the source-view camera parameters.
Each Gaussian in $\Gauss$ is parameterized by a 3D mean $\boldsymbol{\mu}$, a covariance matrix $\boldsymbol{\Sigma}$, an opacity value $\alpha$, and spherical-harmonic color coefficients $\mathbf{h}$. The mean $\boldsymbol{\mu}$ is initialized by back-projecting the corresponding source-view pixel with the refined metric depth $\depthfinal$ and is further adjusted by learned subpixel and depth offsets. The covariance $\boldsymbol{\Sigma}$ is parameterized by anisotropic scale and rotation. The resulting Gaussian set $\Gauss$ is rendered using a differentiable 3D Gaussian splatting renderer~\cite{kerbl20233dgs}.

\subsection{Losses}
\label{sec:method_losses}

Our training objective combines photometric supervision, rendered-depth supervision, and defocus-guided depth supervision.
Let $\Vsup=\{\vs,\vt_1,\ldots,\vt_{T}\}$ denote the set of supervised views, including the source view $\vs$ and the $T$ target views $\{\vt_i\}_{i=1}^{T}$. For each view $v\in\Vsup$, we render an RGB image $\Irender[v]$ and a disparity map $\disprender[v]=1/\depthrender[v]$ using the predicted 3D Gaussians $\Gauss$.
The RGB and depth losses are averaged over all supervised views, whereas the defocus-guided loss is applied only to the source-view U-Net prediction:
\begin{equation}
  \mathcal{L}
  =
  \frac{1}{|\Vsup|}
  \sum_{v\in\Vsup}
  \left(
    \mathcal{L}_{\mathrm{rgb}}^{v}
    +
    \mathcal{L}_{\mathrm{depth}}^{v}
  \right)
  +
  \mathcal{L}_{\mathrm{defocus}}.
\end{equation}

\subsubsection{RGB loss}

For each supervised view $v$, the RGB loss compares the rendered image $\Irender[v]$ with the corresponding ground-truth image $\Iaif[v]$. It combines L1, DSSIM, and LPIPS terms:
\begin{equation}
  \mathcal{L}_{\mathrm{rgb}}^{v} =
  \lambda_{r1}\|\Irender[v]-\Iaif[v]\|_{1}
  + \lambda_{rs}\mathcal{L}_{\mathrm{dssim}}(\Irender[v],\Iaif[v])
  + \lambda_{rp}\mathcal{L}_{\mathrm{lpips}}(\Irender[v],\Iaif[v]),
\end{equation}
where $\mathcal{L}_{\mathrm{dssim}}(x,y)=(1-\mathrm{SSIM}(x,y))/2$, and $\mathcal{L}_{\mathrm{lpips}}$ denotes the LPIPS loss~\cite{zhang2018lpips}.

\subsubsection{Depth loss}

We supervise rendered depth in normalized disparity space. Both the rendered metric disparity $\disprender[v]$ and the ground-truth metric disparity $\disp[v]$ are normalized using the minimum and maximum valid ground-truth disparities of view $v$:
\begin{equation}
  \dispnormrender[v] = \frac{\disprender[v]-\dmin[v]}{\dmax[v]-\dmin[v]},
\end{equation}
\begin{equation}
  \dispnorm[v] = \frac{ \disp[v]-\dmin[v]}{\dmax[v]-\dmin[v]},
\end{equation}
where $\pxvalidset$ denotes the set of pixels with valid ground-truth disparity.
The depth loss combines L1, DSSIM, and gradient terms:
\begin{equation}
  \begin{aligned}
    \mathcal{L}_{\mathrm{depth}}^{v} &=
    \lambda_{d1} \|\dispnormrender[v]-\dispnorm[v]\|_{1}
    + \lambda_{ds} \mathcal{L}_{\mathrm{dssim}}(\dispnormrender[v],\dispnorm[v]) \\
    &\quad + \lambda_{dg} \mathcal{L}_{\mathrm{grad}}(\dispnormrender[v],\dispnorm[v]),
  \end{aligned}
\end{equation}
where
\begin{equation}
  \mathcal{L}_{\mathrm{grad}}(\disprender,\disp) =
  \|\nabla_{x}\disprender-\nabla_{x}\disp\|_{1}
  + \|\nabla_{y}\disprender-\nabla_{y}\disp\|_{1}.
\end{equation}
Here, $\nabla_x$ and $\nabla_y$ denote horizontal and vertical image gradients, respectively.
This term encourages preservation of local disparity structures.

\subsubsection{Defocus-guided depth loss}

To encourage the defocus-conditioned U-Net to predict metric disparity from defocus cues, we supervise its source-view prediction in normalized disparity space. We normalize the U-Net prediction using the source-view ground-truth disparity range:
\begin{equation}
  \dispnormunet = \frac{\dispunet-\dmin[\vs]}{\dmax[\vs]-\dmin[\vs]}.
\end{equation}
The defocus-guided loss is defined as a confidence-weighted MSE:
\begin{equation}
  \mathcal{L}_{\mathrm{defocus}} =
  \lambda_{\mathrm{defocus}}
  \frac{ \sum_{\px \in \pxvalidset}\confunet(\px) \left( \dispnormunet(\px)-\dispnorm[\vs](\px) \right)^{2}}
  { \sum_{\px \in \pxvalidset}\confunet(\px)},
\end{equation}
where $\dispnorm[\vs]$ is the source-view ground-truth normalized disparity and $\confunet$ is the predicted confidence map. The confidence map downweights pixels where the defocus-based estimate is unreliable.

\subsection{Training Data Generation}
\label{sec:trainingdata}

To obtain diverse training data with ground truth at scale, we constructed a synthetic dataset. We expect defocus cues to support synthetic-to-real generalization due to their relative invariance to scene appearance, as suggested by prior work~\cite{maximov2020defocusnet}. The dataset provides all-in-focus images, defocused images, ground-truth depth, and corresponding target views.

The dataset comprises 11\,000 multi-view RGB-D scenes generated using Blender\footnote{\url{http://www.blender.org}}, with 10\,000 scenes for training and 1\,000 for validation. Each scene contained randomly arranged primitive objects with varied materials and lighting. We rendered one source view and four target views per scene. Defocused source-view images were synthesized using the official pretrained BokehMe model~\cite{peng2022bokehme}, with the focus plane set to the nearest scene depth and varying defocus strength. Additional generation details are provided in the supplementary document (\cref{sec:training_data_details}).

%-------------------------------------------------------------------------
\section{Experiments}
\label{sec:experiments}

We compare our method against existing baselines in terms of metric depth estimation accuracy and novel view synthesis (NVS) quality.
Beyond these comparisons, we investigate the effects of f-number and field of view and conduct ablation studies to assess the contributions of individual components.
We further evaluate our method on a real dataset captured with optical defocus, examining reconstructed scene structure and selected physical distances.

\subsection{Datasets}
\label{sec:eval_datasets}

We prepared RGB-D datasets with synthetically generated defocus and a real dataset captured with optical defocus. All input images were resized to $256 \times 256$.
Additional dataset details are provided in the supplementary document (\cref{sec:eval_data_details}).

\subsubsection{RGB-D Datasets with Synthetic Defocus}

We created synthetic-defocus datasets from existing stereo and multi-view datasets with metric depth.
We used Middlebury~\cite{scharstein2014middlebury}, Booster~\cite{ramirez2024booster}, and WildRGB-D~\cite{xia2024wildrgbd}, comprising 12, 38, and 460 scenes, respectively. These datasets provide real images with ground-truth disparity or depth. We generated defocused source-view images using BokehMe~\cite{peng2022bokehme}, setting the focus plane to the minimum valid scene depth. Unless otherwise noted, we used f/4.0 for Middlebury and Booster and f/2.8 for WildRGB-D.

\subsubsection{Real Dataset}

We captured five real-world scenes with optical defocus. All-in-focus--defocused image pairs were acquired from the same viewpoint at focal lengths of 24--34~mm or 70~mm, using f/22 for the all-in-focus images and f/4.0 for the defocused images. Since dense ground-truth depth maps and target-view images were unavailable, we qualitatively assessed reconstructed scene structure and evaluated metric scale using selected physical distances measured in each scene.

\subsection{Implementation Details}

\label{sec:implementation_details}

We implemented our method in PyTorch 2.11.
We used Depth~Anything~3 Large as the frozen monocular prior and trained the defocus-conditioned U-Net $\unetdefocus$, disparity-refinement U-Net $\unetrefine$, and Gaussian parameter prediction module.
The two U-Nets shared the same architecture, with a base channel width of $8$ and $6$ encoder--decoder stages.
We set $\deltadispmax$ to $10.0~\mathrm{m}^{-1}$.
For Gaussian prediction, the RGB image, DA3, and depth tokens were embedded into dimensions of $64$, $256$, and $64$, respectively.
The fused token dimension was $256$, and the decoded dense feature map had $128$ channels.
We used spherical harmonics of degree $2$ for view-dependent color representation.

We optimized all trainable modules using Adam with a base learning rate of $1.5\times10^{-4}$ and a linear warm-up over the first $2\,000$ steps.
Training was run for $100\,000$ steps with a batch size of $4$ and gradient accumulation over $8$ steps, resulting in an effective batch size of $32$.
Gradients were clipped to a norm of $1.0$.
The loss weights were set to $\lambda_{r1}=0.8$, $\lambda_{rs}=0.2$, $\lambda_{rp}=0.1$, $\lambda_{d1}=0.8$, $\lambda_{ds}=0.2$, $\lambda_{dg}=1.0$, and $\lambda_{\mathrm{defocus}}=10.0$.
Training and inference were performed on an NVIDIA GeForce RTX 4080 GPU with 16~GB of memory.

\begin{figure*}
  \centering
  \includegraphics[width=\linewidth]{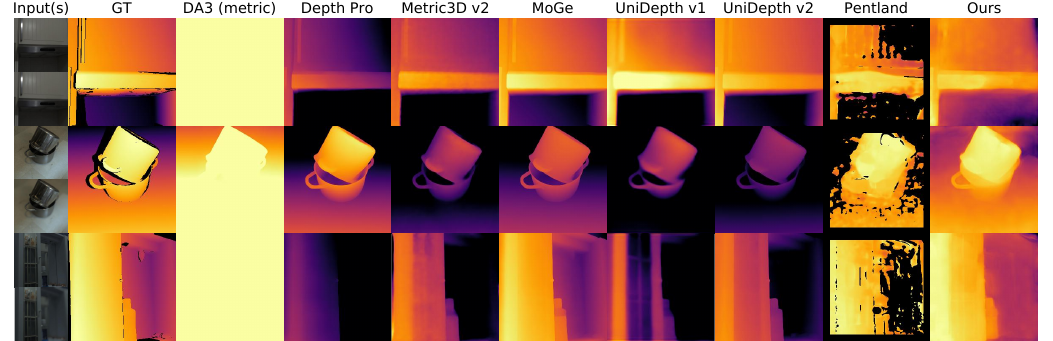}
  \caption{Depth comparisons on the Booster dataset. For better visualization, the brightness of the input images was increased by a factor of 1.2.}
  \label{fig:results_depth_booster}
\end{figure*}

\begin{figure*}
  \centering
  \includegraphics[width=\linewidth]{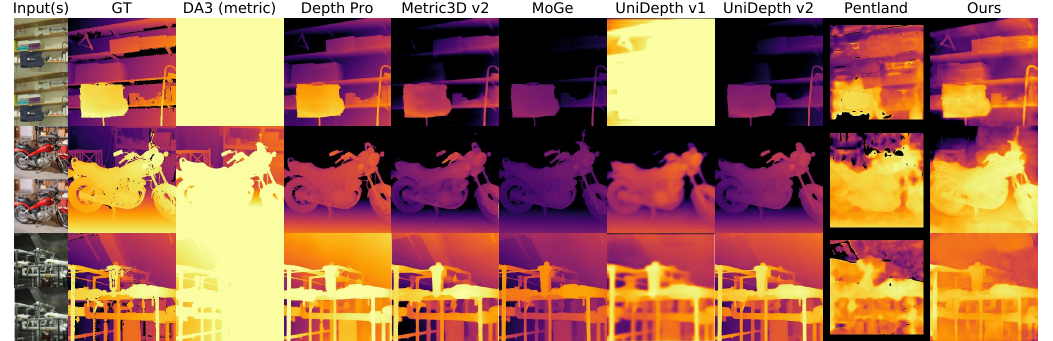}
  \caption{Depth comparisons on the Middlebury dataset.}
  \label{fig:results_depth_middlebury}
\end{figure*}

\begin{figure*}
  \centering
  \includegraphics[width=\linewidth]{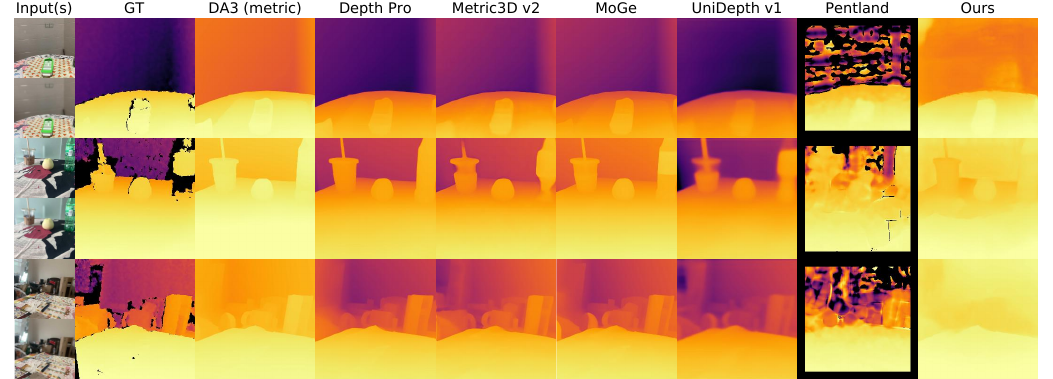}
  \caption{Depth comparisons on the WildRGB-D dataset.}
  \label{fig:results_depth_wildrgbd}
\end{figure*}

\subsection{Metric Depth Evaluation}

\begin{table}[t]
\centering
\caption{Depth comparisons on the Booster dataset.}
\label{tab:results_depth_booster}
\begin{tabular}{lccc}
\toprule
Method & AbsRel$\downarrow$ & RMSE$\downarrow$ & $\delta_1$$\uparrow$ \\
\midrule
DA3~\cite{lin2026da3} (metric) & 0.426 $\pm$ 0.110 & 0.500 $\pm$ 0.268 & 0.06 $\pm$ 0.22 \\
Depth Pro~\cite{bochkovskii2025depthpro} & 0.271 $\pm$ 0.257 & 0.283 $\pm$ 0.220 & 0.63 $\pm$ 0.44 \\
Metric3D v2~\cite{hu2024metric3dv2} & 0.372 $\pm$ 0.302 & 0.392 $\pm$ 0.250 & 0.41 $\pm$ 0.43 \\
MoGe~\cite{wang2025moge} & 0.291 $\pm$ 0.301 & 0.292 $\pm$ 0.265 & 0.51 $\pm$ 0.45 \\
UniDepth v1~\cite{piccinelli2024unidepth} & 0.283 $\pm$ 0.243 & 0.339 $\pm$ 0.257 & 0.62 $\pm$ 0.40 \\
UniDepth v2~\cite{piccinelli2025unidepthv2} & 0.230 $\pm$ 0.200 & 0.248 $\pm$ 0.177 & 0.64 $\pm$ 0.44 \\
Pentland~\cite{pentland1987new} & 0.102 $\pm$ 0.085 & 0.213 $\pm$ 0.230 & 0.83 $\pm$ 0.18 \\
Ours & \cellcolor{teal!20}\textbf{0.075 $\pm$ 0.041} & \cellcolor{teal!20}{0.142 $\pm$ 0.116} & \cellcolor{teal!20}\textbf{0.93 $\pm$ 0.10} \\
\midrule
w/o DA3 Prior & \cellcolor{teal!20}\textbf{0.075 $\pm$ 0.041} & \cellcolor{teal!20}\textbf{0.141 $\pm$ 0.121} & \cellcolor{teal!20}{0.92 $\pm$ 0.12} \\
w/o Scale Map & 0.285 $\pm$ 0.097 & 0.355 $\pm$ 0.119 & 0.57 $\pm$ 0.20 \\
w/o Hidden Gauss & 0.077 $\pm$ 0.042 & 0.152 $\pm$ 0.120 & 0.92 $\pm$ 0.12 \\
\bottomrule
\end{tabular}
\par\smallskip
\begin{minipage}{\linewidth}
\footnotesize\raggedright
Pentland coverage: $52.17 \pm 13.64\%$.
\end{minipage}
\end{table}

\begin{table}[t]
\centering
\caption{Depth comparisons on the Middlebury dataset.}
\label{tab:results_depth_middlebury}
\begin{tabular}{lccc}
\toprule
Method & AbsRel$\downarrow$ & RMSE$\downarrow$ & $\delta_1$$\uparrow$ \\
\midrule
DA3~\cite{lin2026da3} (metric) & 0.430 $\pm$ 0.114 & 1.183 $\pm$ 0.665 & 0.08 $\pm$ 0.18 \\
Depth Pro~\cite{bochkovskii2025depthpro} & 0.237 $\pm$ 0.254 & 0.662 $\pm$ 0.618 & 0.70 $\pm$ 0.40 \\
Metric3D v2~\cite{hu2024metric3dv2} & 0.236 $\pm$ 0.226 & 0.614 $\pm$ 0.509 & 0.80 $\pm$ 0.36 \\
MoGe~\cite{wang2025moge} & 0.315 $\pm$ 0.321 & 0.737 $\pm$ 0.608 & 0.67 $\pm$ 0.41 \\
UniDepth v1~\cite{piccinelli2024unidepth} & 0.189 $\pm$ 0.156 & 0.568 $\pm$ 0.437 & 0.64 $\pm$ 0.41 \\
UniDepth v2~\cite{piccinelli2025unidepthv2} & 0.229 $\pm$ 0.162 & 0.636 $\pm$ 0.461 & 0.54 $\pm$ 0.45 \\
Pentland~\cite{pentland1987new} & \cellcolor{teal!20}\textbf{0.085 $\pm$ 0.036} & \cellcolor{teal!20}\textbf{0.398 $\pm$ 0.350} & 0.89 $\pm$ 0.08 \\
Ours & \cellcolor{teal!20}{0.098 $\pm$ 0.052} & \cellcolor{teal!20}{0.431 $\pm$ 0.422} & \cellcolor{teal!20}\textbf{0.92 $\pm$ 0.11} \\
\midrule
w/o DA3 Prior & 0.881 $\pm$ 2.341 & 4.586 $\pm$ 12.791 & 0.79 $\pm$ 0.18 \\
w/o Scale Map & 0.281 $\pm$ 0.068 & 0.769 $\pm$ 0.268 & 0.48 $\pm$ 0.16 \\
w/o Hidden Gauss & 0.109 $\pm$ 0.066 & 0.475 $\pm$ 0.431 & \cellcolor{teal!20}{0.90 $\pm$ 0.12} \\
\bottomrule
\end{tabular}
\par\smallskip
\begin{minipage}{\linewidth}
\footnotesize\raggedright
Pentland coverage: $67.41 \pm 13.20\%$.
\end{minipage}
\end{table}

\begin{table}[t]
\centering
\caption{Depth comparisons on the WildRGB-D dataset.}
\label{tab:results_depth_wildrgbd}
\begin{tabular}{lccc}
\toprule
Method & AbsRel$\downarrow$ & RMSE$\downarrow$ & $\delta_1$$\uparrow$ \\
\midrule
DA3~\cite{lin2026da3} (metric) & 0.323 $\pm$ 0.101 & 0.740 $\pm$ 0.288 & 0.35 $\pm$ 0.28 \\
Depth Pro~\cite{bochkovskii2025depthpro} & 0.561 $\pm$ 0.427 & 0.448 $\pm$ 0.260 & 0.36 $\pm$ 0.27 \\
Metric3D v2~\cite{hu2024metric3dv2} & 0.514 $\pm$ 0.288 & \cellcolor{teal!20}\textbf{0.387 $\pm$ 0.185} & 0.36 $\pm$ 0.24 \\
MoGe~\cite{wang2025moge} & 0.503 $\pm$ 0.367 & \cellcolor{teal!20}{0.406 $\pm$ 0.217} & 0.39 $\pm$ 0.26 \\
UniDepth v1~\cite{piccinelli2024unidepth} & 0.769 $\pm$ 0.392 & 0.493 $\pm$ 0.188 & 0.29 $\pm$ 0.20 \\
Pentland~\cite{pentland1987new} & 0.735 $\pm$ 1.977 & 115.1 $\pm$ 502.5 & \cellcolor{teal!20}\textbf{0.72 $\pm$ 0.13} \\
% Pentland~\cite{pentland1987new} & 0.735 $\pm$ 1.977 & 115.059 $\pm$ 502.454 & \cellcolor{teal!20}\textbf{0.72 $\pm$ 0.13} \\
Ours & \cellcolor{teal!20}\textbf{0.272 $\pm$ 0.111} & 0.865 $\pm$ 0.395 & \cellcolor{teal!20}{0.59 $\pm$ 0.18} \\
\midrule
w/o DA3 Prior & 0.286 $\pm$ 0.104 & 0.879 $\pm$ 0.408 & 0.57 $\pm$ 0.17 \\
w/o Scale Map & 0.303 $\pm$ 0.117 & 0.756 $\pm$ 0.396 & 0.49 $\pm$ 0.21 \\
w/o Hidden Gauss & \cellcolor{teal!20}{0.286 $\pm$ 0.109} & 0.892 $\pm$ 0.388 & 0.58 $\pm$ 0.18 \\
\bottomrule
\end{tabular}
\par\smallskip
\begin{minipage}{\linewidth}
\footnotesize\raggedright
Pentland coverage: $72.92 \pm 4.64\%$.
\end{minipage}
\end{table}

We compared our method with state-of-the-art single-image metric depth estimation methods, including Depth Pro~\cite{bochkovskii2025depthpro}, Metric3D v2~\cite{hu2024metric3dv2}, MoGe~\cite{wang2025moge}, UniDepth v1~\cite{piccinelli2024unidepth}, UniDepth v2~\cite{piccinelli2025unidepthv2}, and Depth Anything 3 (DA3)~\cite{lin2026da3}. For DA3, we used its metric-depth variant, DA3Metric-Large. Since WildRGB-D is used to train UniDepth v2, we excluded UniDepth v2 from the WildRGB-D evaluation.
We also included Pentland's classical DfD method~\cite{pentland1987new}, using the same all-in-focus--defocused image pair as our method.
We evaluated depth accuracy on the source views using absolute relative error (AbsRel), root mean squared error (RMSE), and threshold accuracy $\delta_1$, defined as the fraction of valid pixels satisfying $\max(\depthrender/\depth, \depth/\depthrender) < 1.25$.
For Pentland, we evaluated only pixels with valid predicted and ground-truth depths. We also report coverage as the fraction of valid ground-truth pixels with valid predictions.
Unless stated otherwise, we report the mean $\pm$ standard deviation over each dataset.

\cref{tab:results_depth_booster,tab:results_depth_middlebury,tab:results_depth_wildrgbd} report quantitative depth comparisons on Booster, Middlebury, and WildRGB-D. Our method outperforms all single-image baselines across the three metrics on Booster and Middlebury. On WildRGB-D, it achieves lower AbsRel and higher $\delta_1$ than these baselines, but has higher RMSE.
Pentland yields lower AbsRel and RMSE on Middlebury and higher $\delta_1$ on WildRGB-D when evaluated on its valid-prediction subsets. However, it produces incomplete depth maps, particularly in weakly textured regions.

\cref{fig:results_depth_booster,fig:results_depth_middlebury,fig:results_depth_wildrgbd} show qualitative depth comparisons. For visualization, each predicted depth map is normalized using the depth range of the corresponding ground-truth depth map. Black regions in the ground-truth depth maps indicate missing depth values, which were masked out during metric calculation.
On Booster and Middlebury, most methods recover the major scene structures but exhibit global scale bias, predicting the entire scene consistently closer or farther than the ground truth. In contrast, our method produces depth maps with a more accurate global metric scale.
On WildRGB-D, our method recovers the coarse scene layout but tends to predict depths that are globally closer than the ground truth. Such errors can increase RMSE even when AbsRel and $\delta_{1}$ improve, because large absolute errors at greater depths are penalized more heavily by RMSE. This behavior may be attributed to weaker defocus cues caused by the smaller focal length and sensor size.

\begin{figure}
  \centering
  \includegraphics[width=.96\linewidth]{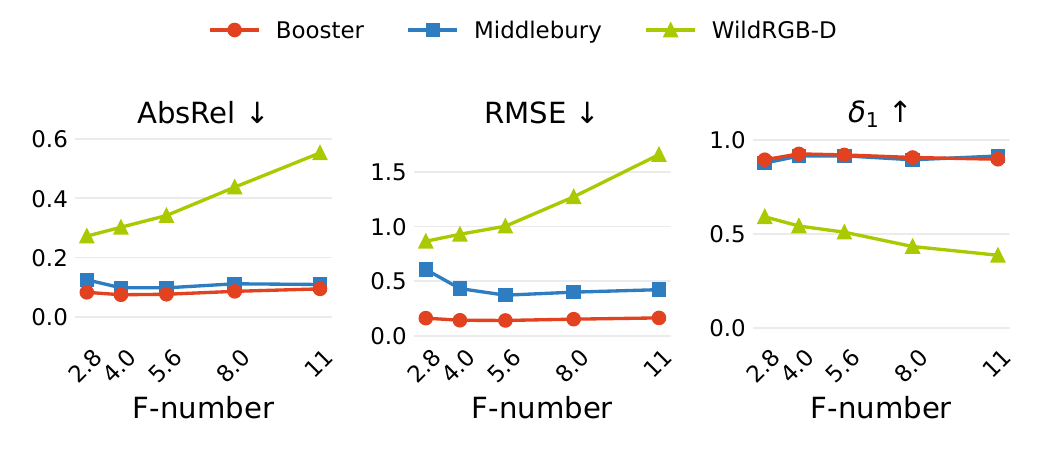}
  \caption{Effect of f-number on depth estimation accuracy.}
  \label{fig:results_f_depth}
\end{figure}

\begin{figure}
  \centering
  \subfloat[Booster dataset.\label{fig:results_crop_depth_booster}]{ \includegraphics[width=.96\linewidth]{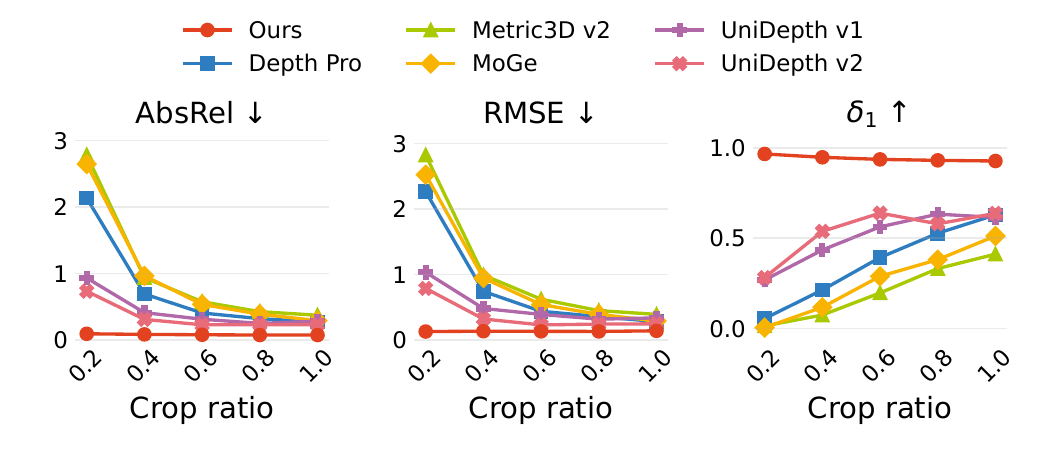} }
  \\
  \subfloat[Middlebury dataset.\label{fig:results_crop_depth_middlebury}]{ \includegraphics[width=.96\linewidth]{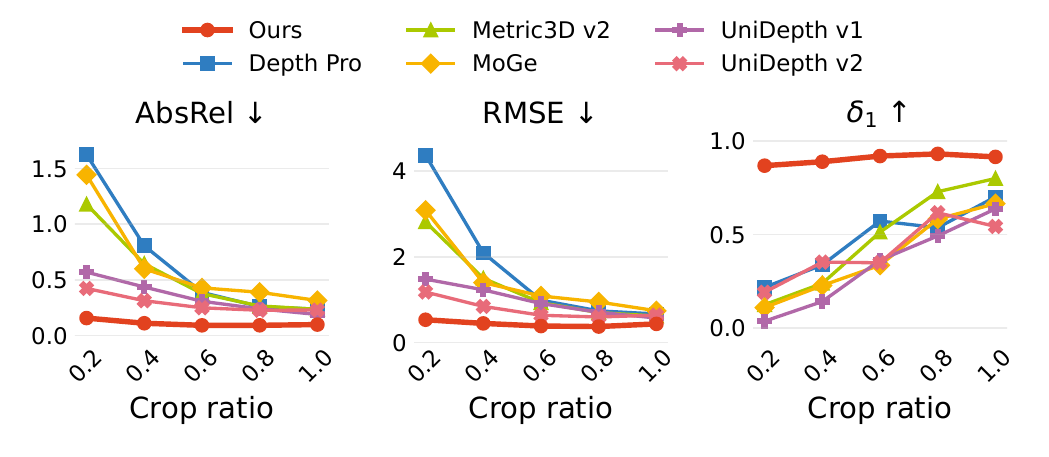} }
  \\
  \caption{Effect of field of view on depth estimation accuracy.}
  \label{fig:results_crop_depth}
\end{figure}

\subsubsection{Effect of f-number}

We analyzed the effect of aperture size by regenerating defocused images $\Idef[\vs]$ with f-numbers of 2.8, 4.0, 5.6, 8.0, and 11 using BokehMe~\cite{peng2022bokehme}.
\cref{fig:results_f_depth} shows the depth estimation accuracy under these f-number settings.
Booster and Middlebury perform best at f/4.0--f/5.6 and remain relatively stable at larger f-numbers. In contrast, WildRGB-D performs best at f/2.8 across all three metrics, with accuracy progressively degrading as the f-number increases.
The stability on Booster and Middlebury suggests that precise aperture tuning is less critical under these imaging conditions. In contrast, the WildRGB-D trend is consistent with weak defocus cues: f/2.8 produces the strongest blur among the tested settings, whereas larger f-numbers further reduce blur and may make depth estimation less reliable.

\subsubsection{Effect of field of view}

We analyzed the effect of field-of-view changes by varying the crop ratio used to generate the input image. Specifically, we center-cropped the all-in-focus images and depth maps at different crop ratios, resized them to $256 \times 256$, and adjusted the camera intrinsics accordingly. We then synthesized defocused images at this resolution, keeping $\Kdef$ fixed across crop ratios for each scene. A smaller crop ratio corresponds to a narrower field of view.
\cref{fig:results_crop_depth} shows the depth estimation accuracy on Booster and Middlebury for crop ratios from 0.2 to 1.0. We did not include WildRGB-D in this analysis because its original image resolution is insufficient for extracting small center crops while maintaining the target resolution of $256 \times 256$.
Compared with the baselines, our method is substantially more stable under changes in crop ratio. This stability can be attributed to the use of defocus as a physical cue. When the crop ratio becomes small, the field of view is narrowed and the semantic and contextual cues available to monocular depth baselines become limited. In contrast, defocus blur is tied to metric depth, allowing our method to maintain a more stable metric scale even under large field-of-view changes.

\subsection{NVS Evaluation}

We compared our method with state-of-the-art single-image novel view synthesis methods, including SHARP~\cite{mescheder2026sharp}, Flash3D~\cite{szymanowicz2025flash3d}, and AdaMPI~\cite{han2022adampi}. SHARP uses a Depth-Pro-like depth prediction backbone~\cite{bochkovskii2025depthpro}. For a fair comparison, we provided the latest Depth Pro predictions to Flash3D and AdaMPI as their input depth maps.
We evaluated rendering quality at the target views using peak signal-to-noise ratio (PSNR), structural similarity index measure (SSIM), and learned perceptual image patch similarity (LPIPS)~\cite{zhang2018lpips}.

\begin{figure*}
  \centering
  \includegraphics[width=\linewidth]{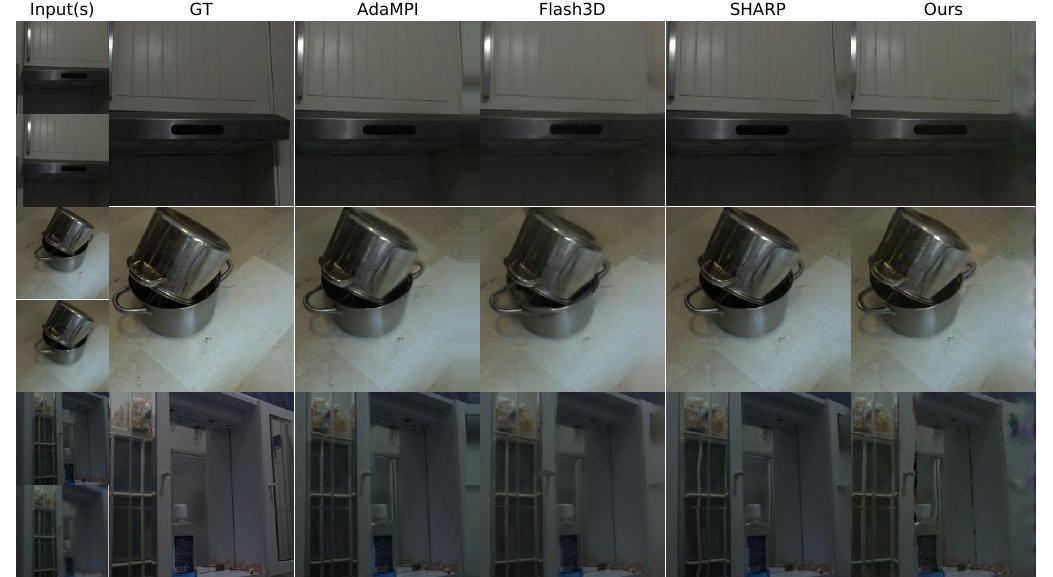}
  \caption{NVS comparisons on the Booster dataset. For better visualization, the image brightness was increased by a factor of 1.2.}
  \label{fig:results_nvs_booster}
\end{figure*}

\begin{figure*}
  \centering
  \includegraphics[width=\linewidth]{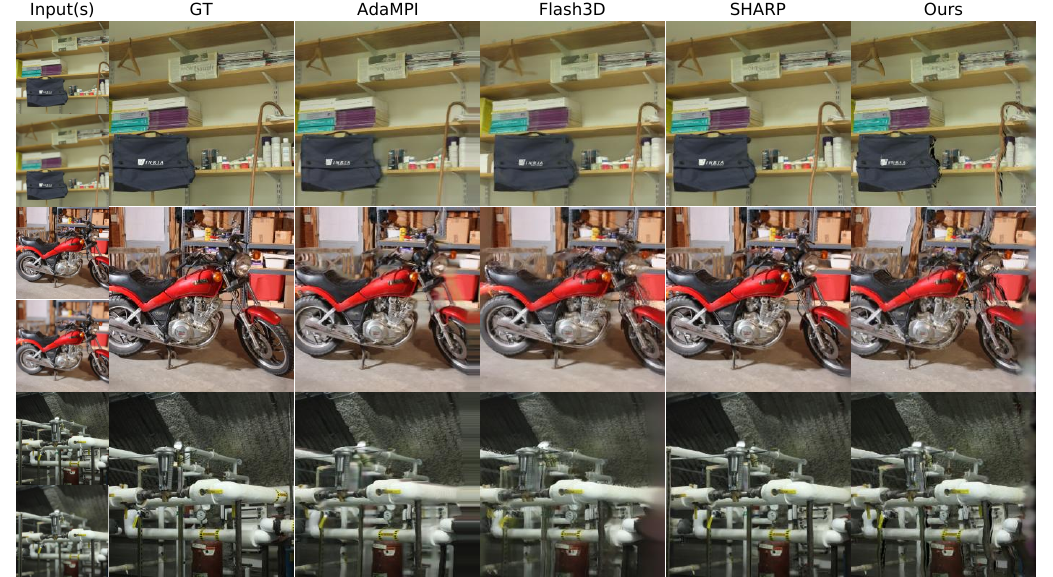}
  \caption{NVS comparisons on the Middlebury dataset.}
  \label{fig:results_nvs_middlebury}
\end{figure*}

\begin{figure*}
  \centering
  \includegraphics[width=\linewidth]{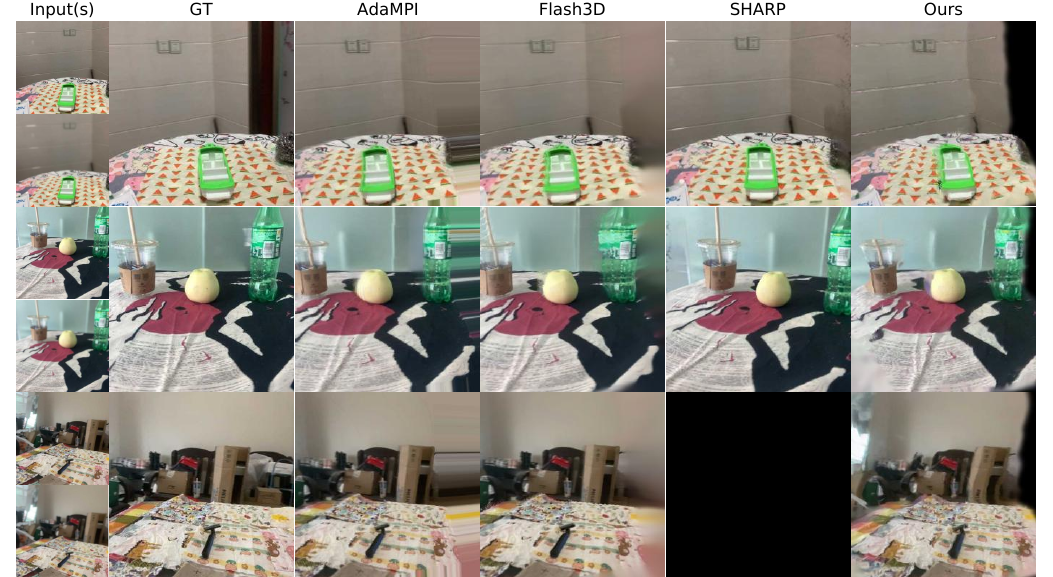}
  \caption{NVS comparisons on the WildRGB-D dataset.}
  \label{fig:results_nvs_wildrgbd}
\end{figure*}

\cref{tab:results_nvs_booster,tab:results_nvs_middlebury,tab:results_nvs_wildrgbd} report quantitative NVS comparisons on Booster, Middlebury, and WildRGB-D, respectively. On Booster and Middlebury, our method outperforms all single-image NVS baselines across all metrics. On WildRGB-D, our method obtains the best LPIPS, while Flash3D achieves the best PSNR and SSIM.
Qualitative comparisons are shown in \cref{fig:results_nvs_booster,fig:results_nvs_middlebury,fig:results_nvs_wildrgbd}. On Booster and Middlebury, our method better preserves the global object layout because the estimated metric scale is more consistent with the ground truth. On WildRGB-D, the foreground regions are well aligned, but the background is estimated to be substantially closer than the ground truth, resulting in larger apparent motion than expected. We also observed geometric distortions in regions with overly smooth depth estimates and color artifacts in disoccluded regions.

\begin{table}[t]
\centering
\caption{NVS comparisons on the Booster dataset.}
\label{tab:results_nvs_booster}
\begin{tabular}{lccc}
\toprule
Method & PSNR$\uparrow$ & SSIM$\uparrow$ & LPIPS$\downarrow$ \\
\midrule
AdaMPI~\cite{han2022adampi} & 22.86 $\pm$ 4.10 & 0.743 $\pm$ 0.137 & 0.285 $\pm$ 0.090 \\
Flash3D~\cite{szymanowicz2025flash3d} & 22.78 $\pm$ 3.94 & 0.737 $\pm$ 0.134 & 0.313 $\pm$ 0.090 \\
SHARP~\cite{mescheder2026sharp} & 22.36 $\pm$ 4.08 & 0.738 $\pm$ 0.135 & \cellcolor{teal!20}{0.262 $\pm$ 0.097} \\
Ours & \cellcolor{teal!20}{24.37 $\pm$ 3.40} & 0.807 $\pm$ 0.086 & \cellcolor{teal!20}\textbf{0.257 $\pm$ 0.056} \\
\midrule
w/o DA3 Prior & 24.31 $\pm$ 3.35 & \cellcolor{teal!20}\textbf{0.816 $\pm$ 0.083} & 0.267 $\pm$ 0.054 \\
w/o Scale Map & 22.03 $\pm$ 3.83 & 0.716 $\pm$ 0.120 & 0.347 $\pm$ 0.072 \\
w/o Hidden Gauss & \cellcolor{teal!20}\textbf{24.57 $\pm$ 3.49} & \cellcolor{teal!20}{0.813 $\pm$ 0.087} & 0.282 $\pm$ 0.059 \\
\bottomrule
\end{tabular}
\end{table}

\begin{table}[t]
\centering
\caption{NVS comparisons on the Middlebury dataset.}
\label{tab:results_nvs_middlebury}
\begin{tabular}{lccc}
\toprule
Method & PSNR$\uparrow$ & SSIM$\uparrow$ & LPIPS$\downarrow$ \\
\midrule
AdaMPI~\cite{han2022adampi} & 17.48 $\pm$ 3.83 & 0.503 $\pm$ 0.174 & 0.331 $\pm$ 0.137 \\
Flash3D~\cite{szymanowicz2025flash3d} & 16.07 $\pm$ 4.01 & 0.425 $\pm$ 0.168 & 0.430 $\pm$ 0.139 \\
SHARP~\cite{mescheder2026sharp} & 16.19 $\pm$ 3.64 & 0.454 $\pm$ 0.164 & 0.344 $\pm$ 0.121 \\
Ours & \cellcolor{teal!20}{18.44 $\pm$ 3.25} & \cellcolor{teal!20}{0.576 $\pm$ 0.153} & \cellcolor{teal!20}\textbf{0.310 $\pm$ 0.104} \\
\midrule
w/o DA3 Prior & 17.56 $\pm$ 3.11 & 0.530 $\pm$ 0.155 & 0.360 $\pm$ 0.108 \\
w/o Scale Map & 15.87 $\pm$ 2.53 & 0.439 $\pm$ 0.120 & 0.422 $\pm$ 0.096 \\
w/o Hidden Gauss & \cellcolor{teal!20}\textbf{18.58 $\pm$ 3.25} & \cellcolor{teal!20}\textbf{0.581 $\pm$ 0.145} & \cellcolor{teal!20}{0.328 $\pm$ 0.102} \\
\bottomrule
\end{tabular}
\end{table}

\begin{table}[t]
\centering
\caption{NVS comparisons on the WildRGB-D dataset.}
\label{tab:results_nvs_wildrgbd}
\begin{tabular}{lccc}
\toprule
Method & PSNR$\uparrow$ & SSIM$\uparrow$ & LPIPS$\downarrow$ \\
\midrule
AdaMPI~\cite{han2022adampi} & \cellcolor{teal!20}{15.18 $\pm$ 2.80} & 0.464 $\pm$ 0.135 & \cellcolor{teal!20}{0.421 $\pm$ 0.098} \\
Flash3D~\cite{szymanowicz2025flash3d} & \cellcolor{teal!20}\textbf{15.30 $\pm$ 2.63} & \cellcolor{teal!20}\textbf{0.472 $\pm$ 0.133} & 0.431 $\pm$ 0.097 \\
SHARP~\cite{mescheder2026sharp} & 7.29 $\pm$ 3.31 & 0.115 $\pm$ 0.140 & 0.674 $\pm$ 0.109 \\
Ours & 14.52 $\pm$ 2.80 & \cellcolor{teal!20}{0.470 $\pm$ 0.143} & \cellcolor{teal!20}\textbf{0.406 $\pm$ 0.084} \\
\midrule
w/o DA3 Prior & 14.48 $\pm$ 2.75 & 0.457 $\pm$ 0.139 & 0.423 $\pm$ 0.084 \\
w/o Scale Map & 13.57 $\pm$ 2.77 & 0.436 $\pm$ 0.134 & 0.451 $\pm$ 0.091 \\
w/o Hidden Gauss & 14.52 $\pm$ 2.76 & 0.461 $\pm$ 0.134 & 0.431 $\pm$ 0.079 \\
\bottomrule
\end{tabular}
\end{table}

\subsection{Ablation}
\label{sec:ablation}

\begin{figure*}
  \centering
  \includegraphics[width=\linewidth]{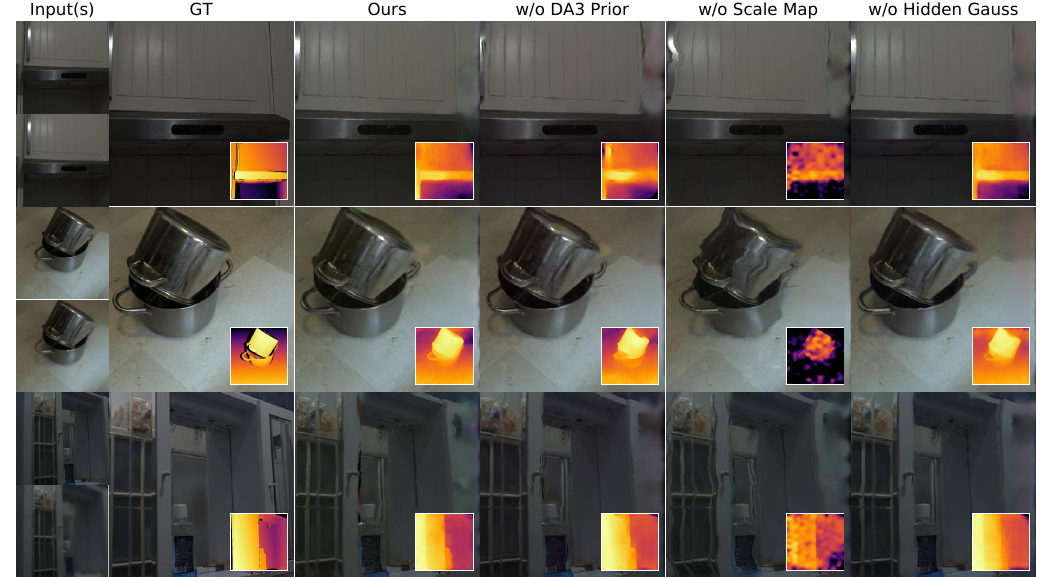}
  \caption{NVS and depth comparisons on the Booster dataset. For better visualization, the image brightness was increased by a factor of 1.2.}
  \label{fig:results_ablation_booster}
\end{figure*}

\begin{figure*}
  \centering
  \includegraphics[width=\linewidth]{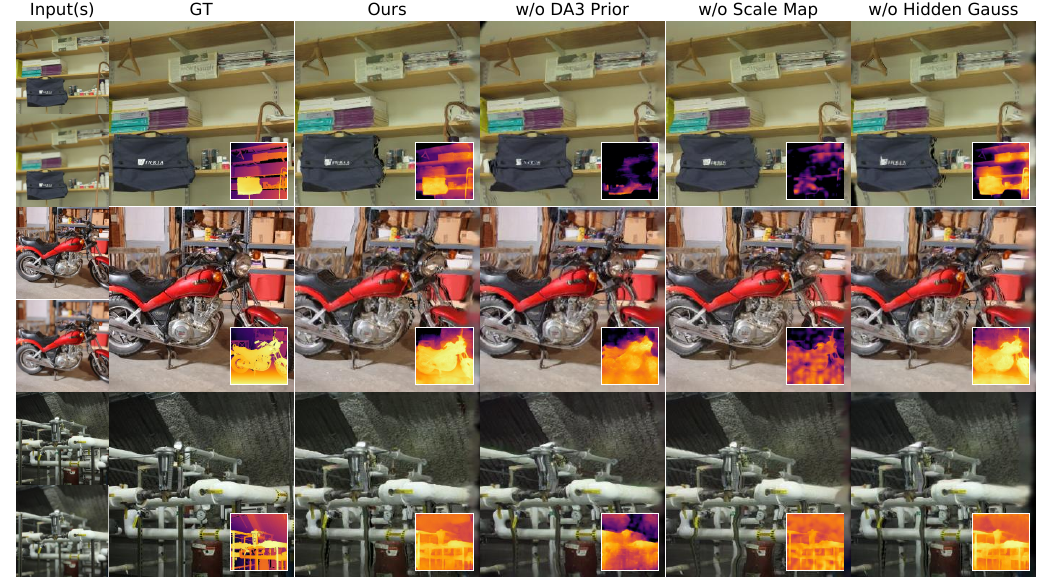}
  \caption{NVS and depth comparisons on the Middlebury dataset.}
  \label{fig:results_ablation_middlebury}
\end{figure*}

\begin{figure*}
  \centering
  \includegraphics[width=\linewidth]{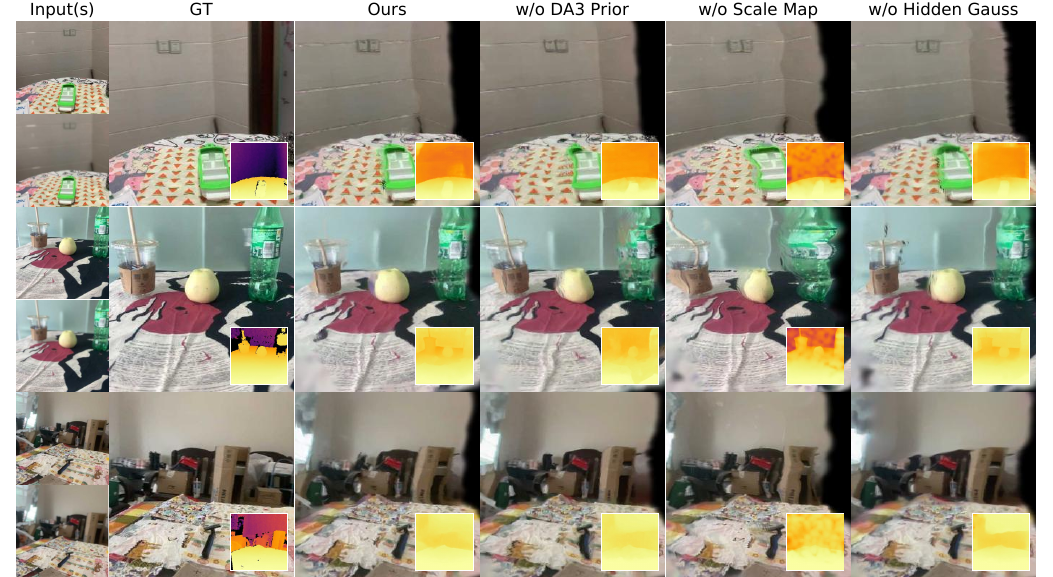}
  \caption{NVS and depth comparisons on the WildRGB-D dataset.}
  \label{fig:results_ablation_wildrgbd}
\end{figure*}

We conducted ablation studies to evaluate the contribution of the DA3 prior, scale-map-based depth refinement, and hidden Gaussian representation.
We compared the full model with three variants. In w/o DA3 prior, we bypass global alignment and dense disparity refinement and directly use the defocus-based U-Net disparity as the final disparity, i.e., $\dispfinal=\dispunet$. In w/o scale map, we remove the dense disparity refinement and set the refined disparity to the globally aligned disparity, i.e., $\dispfinal=\dispalign$. In w/o hidden Gaussian, we predict only the visible Gaussian for each source-view pixel and remove the hidden Gaussian branch.
The quantitative results are reported together with the main comparisons in \cref{tab:results_depth_booster,tab:results_depth_middlebury,tab:results_depth_wildrgbd,tab:results_nvs_booster,tab:results_nvs_middlebury,tab:results_nvs_wildrgbd}, and the qualitative results are shown in \cref{fig:results_ablation_booster,fig:results_ablation_middlebury,fig:results_ablation_wildrgbd}.

Scale-map refinement improves depth accuracy on Booster and Middlebury and all three NVS metrics across datasets. On WildRGB-D, it improves AbsRel and $\delta_{1}$ but increases RMSE. Without refinement, the rendered depth maps show patches of large depth values, while novel views exhibit geometric distortions. Residual local depth errors affect the initial Gaussian positions, and the predictor may attempt to compensate through learned position offsets. These compensatory shifts may disrupt the Gaussian arrangement, contributing to the observed depth irregularities and rendering artifacts. Scale-map refinement reduces local depth errors before Gaussian prediction, limiting the need for such compensation.

The DA3 prior substantially improves depth accuracy on Middlebury, provides smaller gains on WildRGB-D, and has little effect on Booster. It also improves LPIPS across all datasets, although SSIM slightly decreases on Booster. These dataset-dependent depth gains may reflect differences in scene structure. In the Booster examples, defocus-based prediction alone already recovers the main scene geometry. In contrast, the Middlebury examples contain thin and intricate structures whose depth boundaries become overly smoothed without the DA3 prior. These observations suggest that monocular structural guidance and local refinement help preserve fine geometry that is difficult to recover from defocus cues alone.

The hidden Gaussian layer improves depth accuracy and LPIPS across all datasets, although PSNR and SSIM slightly decrease on Booster and Middlebury. Separating visible surfaces from disoccluded content may reduce competing demands on visible Gaussian placement and opacity. However, the hidden layer must predict the colors and textures of surfaces not visible in the input. Even when the reconstructed geometry improves, errors in these predictions may lead to lower PSNR and SSIM.

\subsection{Results on Real Captures}
\label{sec:results_real}

\begin{figure*}
  \centering
  \includegraphics[width=\linewidth]{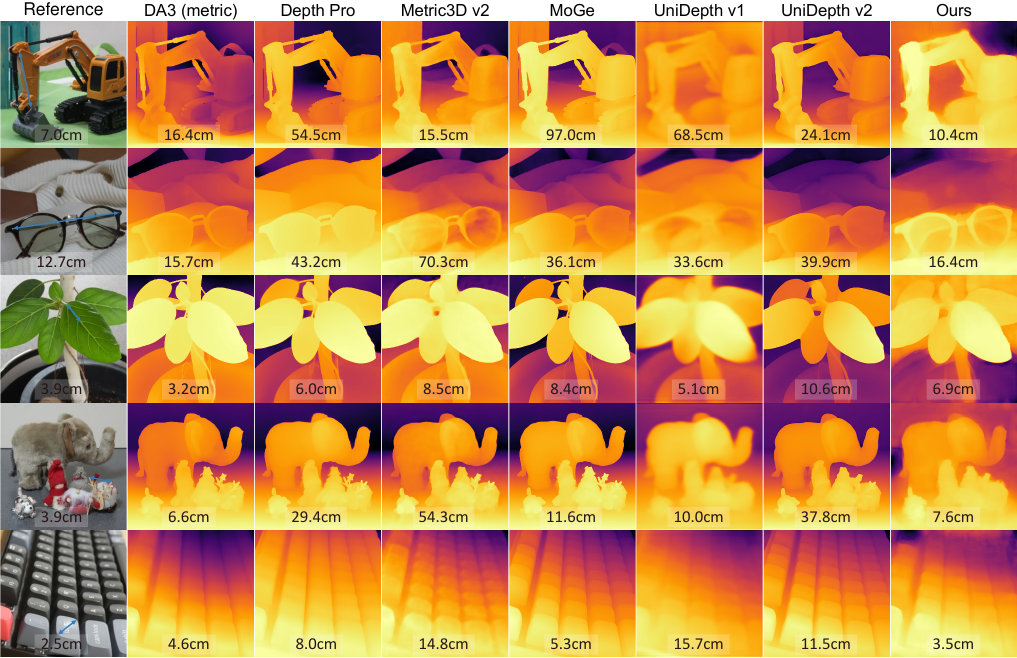}
  \caption{Metric depth comparisons on five real scenes captured at 70~mm. The reference column shows the input images and measured distances. The remaining columns show estimated depth maps (normalized for visualization) and the corresponding estimated distances.}
  \label{fig:results_real_comp}
\end{figure*}

\begin{figure*}
  \centering
  \includegraphics[width=\linewidth]{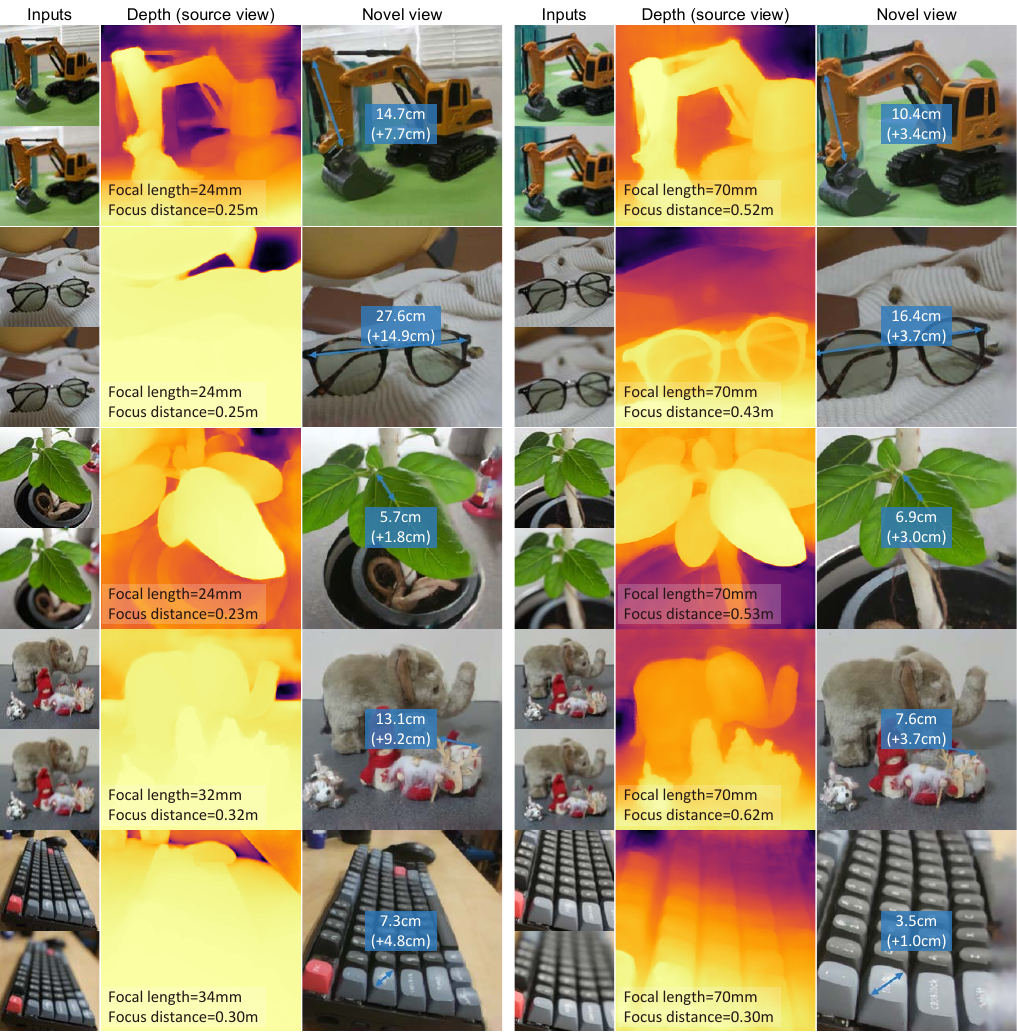}
  \caption{Results on five real scenes captured at shorter focal lengths (24--34~mm, left) and 70~mm (right). Each group shows the all-in-focus--defocused input pair, source-view rendered depth (normalized for visualization), and a novel view. Blue annotations indicate estimated distances and signed errors (estimate minus ground truth). Focal lengths and focus distances are shown in each row.}
  \label{fig:results_real_fl}
\end{figure*}

We evaluated our method on the real dataset comprising five scenes described in \cref{sec:eval_datasets}. We compared estimated depth maps and selected physical distances across methods and examined our results under different capture configurations.

\cref{fig:results_real_comp} compares predicted depth maps and selected physical distances at 70~mm. The depth maps are normalized for visualization, while the distances are reported in metric units.
Our method recovers the main scene structure under real optical defocus, but produces smoother object boundaries than most single-image baselines, particularly around the toy excavator and small decorative objects. Its absolute distance errors are smaller than those of most baselines. DA3 is the strongest single-image baseline, with smaller errors than ours in the eyeglasses, plant, and decorative-object scenes.
\cref{fig:results_real_fl} compares our results under the two capture configurations. The 70~mm captures yield smaller absolute distance errors in four of the five scenes, except for the plant scene.

The remaining errors may reflect both limited defocus sensitivity and differences between the synthetic training data and real captures. Near the shortest focus distance of the 70~mm captures (0.3~m), a one-pixel change in blur radius corresponds locally to approximately 1.4~cm of depth variation under our blur model. This local depth variation grows quadratically with depth (\cref{sec:depth_sensitivity}). Differences in scene content and lens-specific blur may also contribute to the errors. The present experiments do not isolate these factors. Further evaluation with dense ground-truth depth would help characterize their effects.

%-------------------------------------------------------------------------
\section{Conclusion}
\label{sec:conclusion}

We presented DfD-GS, a feed-forward framework that predicts metric-scale 3D Gaussians from a single-view all-in-focus--defocused image pair.
Experiments on synthetic-defocus datasets showed that DfD-GS outperformed the single-image baselines in both tasks on Booster and Middlebury. On WildRGB-D, it achieved lower AbsRel and LPIPS but higher depth RMSE than these baselines. Additional analyses showed more stable metric scale under field-of-view changes and dataset-dependent sensitivity to f-number. Experiments with optical defocus demonstrated recovery of the main scene structure.

Despite these improvements, several limitations remain. The nearest-focus assumption restricts capture conditions. Extending the formulation to arbitrary focus positions is a direction for future work. Residual depth errors and uncertainty about unobserved surfaces limit novel view rendering quality. Future work should better preserve reliable monocular geometry during refinement and improve hidden-Gaussian prediction. Errors in real captures may arise from limited defocus sensitivity and differences between synthetic and optical blur. Evaluation with dense ground-truth depth would help characterize these limitations and guide further improvements.

% Flush all pending figures and tables before the references.
\clearpage
\bibliographystyle{IEEEtran}
\bibliography{references}

% Supplementary material, included by the root main.tex.
\clearpage

% Supplementary-only notation
\newcommand{\focallength}{f}
\newcommand{\fnumber}{N}
\newcommand{\sensorwidth}{s_w}
\newcommand{\imagewidth}{i_w}
\newcommand{\dispmin}{d_{\min}}

% Continue page numbering; use distinct counters and PDF anchors for the supplement.
\setcounter{section}{0}
\setcounter{figure}{0}
\setcounter{table}{0}
\setcounter{equation}{0}
\renewcommand{\thesection}{S\arabic{section}}
\renewcommand{\thefigure}{S\arabic{figure}}
\renewcommand{\thetable}{S\arabic{table}}
\renewcommand{\theequation}{S\arabic{equation}}
\renewcommand{\theHsection}{supp.\arabic{section}}
\renewcommand{\theHfigure}{supp.\arabic{figure}}
\renewcommand{\theHtable}{supp.\arabic{table}}
\renewcommand{\theHequation}{supp.\arabic{equation}}

\twocolumn[{%
    \centering
    {\Large\bfseries \papertitle\par}
    \vspace{1em}
    {\large\bfseries Supplementary Material\par}
    \vspace{1.5em}
}]

%-------------------------------------------------------------------------
\noindent
This supplementary document provides additional details and results as follows:
\begin{itemize}
  \item \textbf{Metric Depth Prediction}: implementation details for computing defocus strength and constraining disparity values.
  \item \textbf{Training Data}: details of synthetic multi-view RGB-D scene generation and defocus synthesis.
  \item \textbf{Evaluation Data}: details of synthetic-defocus datasets and a real dataset.
  \item \textbf{Computational Performance}: prediction and rendering times and GPU memory usage.
  \item \textbf{Training Stability}: performance variation across five training seeds.
  \item \textbf{Depth Sensitivity of Defocus Blur}: theoretical analysis of how scene depth and camera settings affect depth sensitivity.
\end{itemize}

%-------------------------------------------------------------------------
\section{Metric Depth Prediction}
\label{sec:metric_depth_details}

\subsection{Defocus Strength}
\label{sec:defocus_strength}

We compute the defocus strength used in \cref{eq:defocus_relation} as
\begin{equation}
  \Kdef
  = \frac{\focallength^2}{2\fnumber}\,
  \frac{\imagewidth}{\sensorwidth},
  \label{eq:defocus_strength}
\end{equation}
where $\focallength$ and $\sensorwidth$ are the focal length and sensor width in meters, $\fnumber$ is the f-number, and $\imagewidth$ is the image width in pixels. The factor of $1/2$ converts aperture diameter to radius.
The resulting $\Kdef$ has units of $\mathrm{pixel}\cdot\mathrm{m}$, consistent with disparity in $\mathrm{m}^{-1}$ and blur radius in pixels.

\subsection{Disparity Flooring and Clipping}
\label{sec:disparity_clipping}

We use $\dispmin=0.01~\mathrm{m}^{-1}$ as the lower bound on metric disparity. For defocus-guided depth supervision, we obtain $\dispunet$ by applying a smooth positive floor to $\tilde{\dispunet}=\dispfocus-\deltadispunet$:
\begin{equation}
  \dispunet(\px)
  =
  \dispmin
  +
  \tau\log\!\left(
    1+\exp\!\left(
      \frac{\tilde{\dispunet}(\px)-\dispmin}{\tau}
    \right)
  \right),
  \label{eq:unet_metric_disp}
\end{equation}
where $\tau=0.02~\mathrm{m}^{-1}$ controls the smoothness of the floor.

For global alignment and dense refinement, we apply hard clipping:
\begin{align}
  \dispalign(\px)
  &=
  \operatorname{clip}_{[\dispmin,\dispfocus]}
  \!\left(
    \dispfocus-\deltadispalign(\px)
  \right),
  \label{eq:aligned_metric_disparity} \\
  \dispfinal(\px)
  &=
  \operatorname{clip}_{[\dispmin,\dispfocus]}
  \!\left(
    \dispalign(\px)
    \exp\!\left(\scalerange\normscalemap(\px)\right)
  \right).
  \label{eq:refined_metric_disparity}
\end{align}
The lower bound ensures positive disparity, while the upper bound enforces the nearest-focus assumption.

%-------------------------------------------------------------------------
\section{Training Data}
\label{sec:training_data_details}

Each synthetic scene contained 30 primitive objects, including cubes, spheres, cylinders, cones, and tori, randomly distributed within the camera frustum at depths between 0.1 and 10~m. Object materials and lighting conditions were also randomly varied.

The camera was configured with a 55~mm focal length and a $24 \times 24$~mm sensor size. We rendered one source view $(\Iaif[\vs], \depth[\vs])$ and four target views $\{(\Iaif[\vt_i], \depth[\vt_i])\}_{i=1}^{4}$ from horizontally and vertically shifted camera positions. All color and depth maps were rendered at $256 \times 256$ resolution.

We synthesized the defocused source-view image $\Idef[\vs]$ from the all-in-focus image $\Iaif[\vs]$, ground-truth depth $\depth[\vs]$, and defocus strength $\Kdef[\vs]$ using the official pretrained BokehMe model~\cite{peng2022bokehme}. The focus distance was set to the minimum scene depth, corresponding to the maximum scene disparity $\dispfocus$. We sampled $\Kdef[\vs]$ from $[0.3, 10.0]$, with larger values producing stronger defocus blur.

%-------------------------------------------------------------------------
\section{Evaluation Data}
\label{sec:eval_data_details}

\subsection{RGB-D Datasets with Synthetic Defocus}

We created synthetic-defocus datasets from Middlebury~\cite{scharstein2014middlebury} and Booster~\cite{ramirez2024booster}, which provide stereo images with ground-truth disparity, and WildRGB-D~\cite{xia2024wildrgbd}, which provides multi-view RGB-D sequences. For Middlebury and Booster, we used the official training splits, retaining 12 of the 15 Middlebury scenes after excluding those with inconsistent lighting between stereo images and all 38 Booster scenes. For WildRGB-D, we used the validation split and randomly selected up to 10 scenes from each of the 46 categories. The resulting evaluation sets contained 12, 38, and 460 scenes from Middlebury, Booster, and WildRGB-D, respectively.

For Middlebury and Booster, we used the left and right images as the source and target views, respectively, and converted the ground-truth disparity maps into metric depth using the provided calibration parameters. For WildRGB-D, we sampled two frames from each sequence with a fixed temporal stride of 10 frames, using the earlier frame as the source view and the later frame as the target view. Depth estimation was evaluated against the source-view ground truth, while NVS was evaluated by comparing the synthesized target-view image with the corresponding ground-truth image. All images and depth maps were center-cropped to a square aspect ratio and resized to $256 \times 256$.

We synthesized defocused source-view images from the all-in-focus RGB images using BokehMe~\cite{peng2022bokehme}. The focus distance was set to the minimum valid scene depth, placing the closest visible surface in focus. We assumed sensor sizes of $22.2 \times 14.8$~mm, $14.2 \times 10.4$~mm, and $7.5 \times 10.0$~mm for Middlebury, Booster, and WildRGB-D, respectively. Unless otherwise noted, we used f/4.0 for Middlebury and Booster and f/2.8 for WildRGB-D. To obtain the dense depth maps required by BokehMe, we filled missing depth regions using Any2Full~\cite{zhou2026any2full}. The completed depth maps were used only for blur synthesis. Depth evaluation used the original valid ground-truth pixels.

\subsection{Real Dataset}

We captured five real-world scenes using a Sony $\alpha$7C $\rm{I\hspace{-.01em}I}$ camera equipped with an FE 24--70mm F2.8 GM $\rm{I\hspace{-.01em}I}$ lens. For each scene, we acquired all-in-focus--defocused image pairs at focal lengths of 24--34~mm and 70~mm. Each pair was captured from the same viewpoint with the camera mounted on a tripod, using f/22 for the all-in-focus image and f/4.0 for the defocused image. Exposure settings were adjusted to maintain comparable brightness between the two images. All images were captured at $4672 \times 4672$ pixels and resized to $256 \times 256$ before inference.

Since dense ground-truth depth maps and target-view images were unavailable, we qualitatively assessed reconstructed scene structure and evaluated metric scale using selected physical distances measured in each scene. For each measurement, we backprojected the annotated endpoints using the estimated depth and camera intrinsics and computed the Euclidean distance between the resulting 3D points. For our method, the depth maps were rendered from the predicted 3D Gaussians at the source view.

%-------------------------------------------------------------------------
\section{Computational Performance}
\label{sec:computational_performance}

We measured performance on Booster, Middlebury, and WildRGB-D using a single NVIDIA GeForce RTX 4080 GPU at $256 \times 256$ resolution.
\cref{tab:computational_performance} reports the median time to predict the Gaussian representation from the input images and the median time to render each novel view, measured with CUDA synchronization. Peak allocated GPU memory during evaluation is 1.75~GiB for all datasets.

\begin{table}[t]
  \caption{Median Gaussian prediction and per-view rendering times.}
  \label{tab:computational_performance}
  \centering
  \begin{tabular}{lrr}
    \toprule
    Dataset & Prediction (ms) & Rendering (ms) \\
    \midrule
    Booster    & 42.23 & 2.04 \\
    Middlebury & 43.29 & 2.05 \\
    WildRGB-D  & 42.25 & 2.10 \\
    \bottomrule
  \end{tabular}
\end{table}

%-------------------------------------------------------------------------
\section{Training Stability}
\label{sec:training_stability}

We trained five models with different seeds, keeping all other training settings fixed.
\cref{tab:training_stability} reports the mean and standard deviation across the five models, after averaging scenes within each model.

\begin{table}[t]
  \caption{Mean $\pm$ standard deviation across five training seeds.}
  \label{tab:training_stability}
  \centering
  \begin{tabular}{lccc}
    \toprule
    Dataset & AbsRel $\downarrow$ & RMSE $\downarrow$ & $\delta_1$ $\uparrow$ \\
    \midrule
    Booster & $0.084 \pm 0.008$ & $0.156 \pm 0.009$ & $0.91 \pm 0.01$ \\
    Middlebury & $0.114 \pm 0.013$ & $0.467 \pm 0.044$ & $0.90 \pm 0.02$ \\
    WildRGB-D & $0.313 \pm 0.048$ & $1.011 \pm 0.258$ & $0.56 \pm 0.03$ \\
    \midrule
    Dataset & PSNR (dB) $\uparrow$ & SSIM $\uparrow$ & LPIPS $\downarrow$ \\
    \midrule
    Booster & $24.10 \pm 0.16$ & $0.802 \pm 0.008$ & $0.272 \pm 0.016$ \\
    Middlebury & $18.41 \pm 0.24$ & $0.579 \pm 0.011$ & $0.317 \pm 0.008$ \\
    WildRGB-D & $14.38 \pm 0.11$ & $0.452 \pm 0.016$ & $0.423 \pm 0.018$ \\
    \bottomrule
  \end{tabular}
\end{table}

%-------------------------------------------------------------------------
\section{Depth Sensitivity of Defocus Blur}
\label{sec:depth_sensitivity}

The sensitivity of defocus blur to changes in scene depth follows from \cref{eq:defocus_relation,eq:defocus_strength}. Let $z=\depth(\px)$ denote the depth at a pixel. With camera parameters and focus distance fixed, the magnitude of the blur-radius derivative away from the focus plane is
\begin{equation}
  \left|\frac{\partial \coc}{\partial z}\right|
  = \frac{\Kdef}{z^2}.
  \label{eq:coc_depth_derivative}
\end{equation}
Thus, depth sensitivity decreases quadratically with scene depth, while camera parameters and image sampling affect it through $\Kdef$.

For our 70~mm captures at f/4.0, the effective sensor width of 24~mm and input width of 256 pixels give $\Kdef\approx6.53~\mathrm{pixel}\cdot\mathrm{m}$. Near the shortest focus distance in these captures (0.3~m), a one-pixel change in blur radius corresponds locally to approximately 1.4~cm of depth variation.

\end{document}